\PassOptionsToPackage{colorlinks=true, citecolor=blue}{hyperref}
\PassOptionsToPackage{normalem}{ulem}
\PassOptionsToPackage{table}{xcolor}
\documentclass{article}

\usepackage[nonatbib, final]{neurips_2020}

\usepackage[utf8]{inputenc} 
\usepackage[T1]{fontenc}    
\usepackage{hyperref}       
\usepackage{url}            
\usepackage{booktabs}       
\usepackage{amsfonts}       
\usepackage{nicefrac}       
\usepackage{microtype}      
\usepackage{color}

\usepackage{multirow}

\usepackage[style=authoryear, natbib=true, sorting=nyt, maxbibnames=9, maxcitenames=2, uniquelist=false, backend=biber]{biblatex}
\usepackage[subrefformat=parens,labelformat=parens]{subcaption}

\usepackage{mathtools}
\usepackage{bm}
\usepackage{bbm}

\usepackage{physics}

\usepackage{algorithm}
\usepackage{algpseudocode}

\usepackage{dirtytalk}

\usepackage{ulem,xpatch}

\xpatchcmd{\sout}
  {\bgroup}
  {\bgroup}
  {}{}

\usepackage{tikz}
\usetikzlibrary{arrows.meta}

\usepackage[utf8]{inputenc}
\usepackage{pgfplots}
\DeclareUnicodeCharacter{2212}{−}
\usepgfplotslibrary{groupplots,dateplot}
\usetikzlibrary{patterns,shapes.arrows}
\pgfplotsset{compat=newest}

\pgfplotsset{every axis/.style={scale only axis}}

\usepackage{appendix}
\renewcommand{\appendixname}{Supplementary Material}

\renewcommand{\cite}{\citep}

\title{Similarity of Classification Tasks}

\author{%
    Cuong Nguyen\\
    Australian Institute for Machine Learning\\
    University of Adelaide, Australia\\
    \texttt{cuong.nguyen@adelaide.edu.au} \\
    \And
    Thanh-Toan Do\thanks{Work done while at the University of Liverpool, UK}\\
    Faculty of Information Technology\\
    Monash University, Australia\\
    \texttt{toan.do@monash.edu} \\
    \AND
    Gustavo Carneiro \\
    Australian Institute for Machine Learning\\
    University of Adelaide, Australia\\
    \texttt{gustavo.carneiro@adelaide.edu.au}
}

\begin{document}

    \maketitle
    \begin{abstract}
    Recent advances in meta-learning has led to remarkable performances on several few-shot learning benchmarks. However, such success often ignores the similarity between training and testing tasks, resulting in a potential bias evaluation. We, therefore, propose a generative approach based on a variant of Latent Dirichlet Allocation to analyse task similarity to optimise and better understand the performance of meta-learning. We demonstrate that the proposed method can provide an insightful evaluation for meta-learning algorithms on two few-shot classification benchmarks that matches common intuition: the more similar the higher performance. Based on this similarity measure, we propose a task-selection strategy for meta-learning and show that it can produce more accurate classification results than methods that randomly select training tasks.
\end{abstract}
    \section{Introduction}
\label{sec:introduction}

    The vast development in machine learning has enabled possibilities to solve increasingly complex applications. Such complexity require high capacity models, which in turn need a massive amount of annotated data for training, resulting in an arduous, costly and even infeasible annotation process. This has, therefore, motivated the research of novel learning approaches, generally known as transfer learning, that exploit past experience (in the form of models learned from other training tasks) to quickly learn a new task using relatively small training sets.
    
    Transfer-learning, and in particular, meta-learning, has recently achieved state-of-the-art results in several few-shot learning benchmarks~\cite{vinyals2016matching,snell2017prototypical,finn2017model,yoon2018bayesian,rusu2019meta}. Such success depends not only on the effectiveness of transfer learning algorithms, but also on the similarity between training and testing tasks~\cite{chen2020new}. More specifically, the larger the subset of training tasks that are similar to the testing tasks, the higher the classification accuracy on those testing tasks. However, meta-learning methods are assessed without taking into account such observation, which can bias the meta-learning classification results depending on the policy for selecting training and testing tasks.
    
    In this paper, we propose a generative approach based on a \say{continuous} version of Latent Dirichlet Co-Clustering to model classification tasks. The resulting model represents tasks in a latent \say{task-theme} simplex, and hence, allows to quantitatively measure their similarity. The proposed similarity measure enables the possibility of selecting the most related tasks from the training set for the meta-learning of a novel testing task. We empirically demonstrate that the proposed task selection strategy outperforms the one that randomly selects training tasks across several meta-learning methods.
    \section{Related work}
\label{sec:related_work}
    With this paper, we target an improved understanding of meta-learning algorithms~\cite{chen2019closer,dhillon2019baseline}, which can allow us to improve their current performance.
    Although meta-learning has progressed steadily with many remarkable achievements, it has been reported that there is a large variance of performance among testing tasks~\cite{dhillon2019baseline}. This observation suggests that not all testing tasks are equally related to training tasks.
    \say{Task hardness} which is based on the cosine similarity between the embedding of labelled and unlabelled data is, therefore, proposed to better justify the performance of meta-learning methods~\cite{dhillon2019baseline}. This, however, quantifies only the similarity between samples within a task without investigating the similarity between tasks.
    
    Task similarity has been intensively studied in the field of multi-task learning. Some remarkable works include task-clustering using k-nearest neighbours~\cite{thrun1996discovering}, modelling common prior between tasks as a mixture of distributions~\cite{bakker2003task} with the extension using Dirichlet Process~\cite{xue2007multi}, applying a convex formulation to either cluster~\cite{jacob2009clustered} or learning task relationship through task covariance matrices~\cite{zhang2012convex}. Other approaches try to provide theoretical guarantees when learning the similarity or relationship between tasks~\cite{shui2019a}. Following a similar approach, an extensive experiment was carried out on 26 computer-vision tasks to determine the correlation between those tasks, also known as \say{taskonomy}~\cite{zamir2018taskonomy}. Some recent works~\cite{tran2019transferability,nguyen2020leep} take a slightly different approach by investigating the correlation of the label distribution between those tasks of interest. One commonality of those studies is their reliance on a discriminative approach, where the similarity of task-specific classifiers are used to quantify task relatedness. In addition, most of those works focus more on the conventional machine learning setting, which requires a sufficient number of labelled data on the novel tasks to perform transfer learning. In contrast, our proposal follows a generative approach which does not depend on any task-specific classifier. Our approach can also work in the few-shot setting, where only a few labelled data points from the targeted tasks are available. Another work that is slightly related to task similarity is Task2Vec~\cite{achille2019task2vec}, which employs Fisher information matrix of an external network, known as \say{probe} network, to model visual tasks as fixed vectors in an embedding space, allowing to analyse and calculate task similarity. However, its application is still limited due to the need of an external network pre-trained to perform specific tasks on some standard visual data sets.
    
    Our work is also related to finite mixture models~\cite{pritchard2000inference}, such as the Latent Dirichlet Allocation (LDA)~\cite{blei2003latent}, in topic modelling which analyses and summarises text data, or in computer vision~\cite{li2005bayesian}. LDA assumes that each document within a given corpus can be represented as a finite mixture model, where its components are the latent topics shared across all documents. Training an LDA model or its variants on a large text corpus is challenging, so several approximate inference techniques have been proposed, ranging from mean-field variational inference (VI)~\cite{blei2003latent}, collapsed Gibbs' sampling~\cite{griffiths2004finding} and collapsed VI~\cite{teh2007collapsed}. Furthermore, several online inference methods have been developed to increase the training efficiency for large corpora~\cite{canini2009online,hoffman2010online,foulds2013stochastic}. Our work is slightly different from the inference for conventional LDA models, where we perform online learning for Latent Dirichlet Co-clustering~\cite{shafiei2006latent} -- a variant of LDA -- that includes the information of paragraphs into the model. In addition, our approach considers \say{word} as continuous data, instead of the discrete data represented by a bag-of-word vector generally used in topic modelling.
    \section{Method}
\label{sec:methodology}
    	
    	\begin{figure*}[t]
    		\centering
    		\begin{tikzpicture}[scale=0.875, rounded corners=2pt, every node/.style={scale=0.875, minimum size=1.1cm}]
    			\input{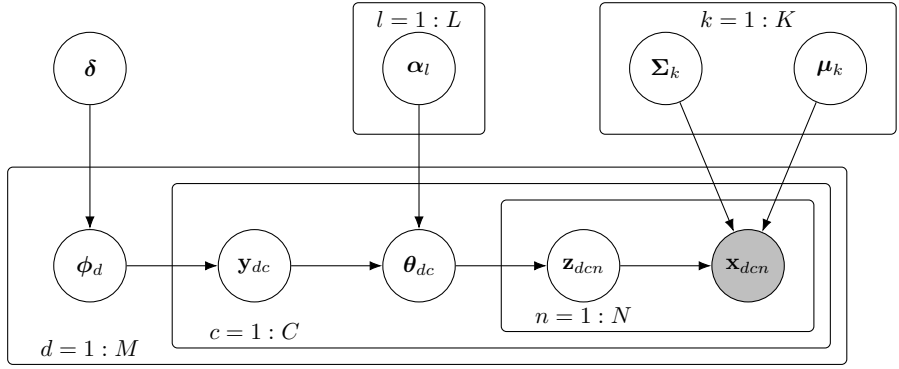}
    		\end{tikzpicture}
    		\caption{Directed acyclic graph represents the continuous LDCC that models classification tasks as a finite mixture of Gaussian distributions.}
    		\label{fig:cclda}
    	\end{figure*}
    	
        To relate image \textbf{classification} to \textbf{topic modelling}, we consider \textbf{a task} as \textbf{a document}, \textbf{a class} as \textbf{a paragraph}, and \textbf{an image} as \textbf{a word}. Given these analogies, we employ the Latent Dirichlet Co-clustering (LDCC)~\cite{shafiei2006latent} -- a variant of LDA -- to model classification tasks. The LDCC extends the conventional LDA to a hierarchical structure by including the information about paragraphs, or in our case, data classes, into the model.
        Since the data in classification is assumed to be continuous, the categorical word-topic distribution in the original LDCC model is replaced by a Gaussian \textbf{image-theme} distribution. Each classification task can be modelled as a mixture of \(L\) \textbf{task-themes} (corresponding to document topic in LDCC), where each task-theme is a \say{summary} of many finite mixtures of \(K\) \textbf{image-themes}. We can, therefore, utilise this representation, and in particular the \textit{task-theme mixture} parameter to quantify the similarity between tasks.
        
        We assume that there are \(M\) classification tasks, where each task consists of \(C\) classes, and each class has \(N\) images (i.e., using meta-learning nomenclature, this represents \(M\) \(C\)-way \(N\)-shot classification tasks). For simplicity, \(C\) and \(N\) are assumed to be fixed across all tasks, but the extension of varying \(C\) and \(N\) is trivial and can be implemented straightforwardly. The process to generate classification tasks from an \say{\(L\)-task-\(K\)-image theme} model shown in \figureautorefname{~\ref{fig:cclda}} can be presented as follows:
    	\begin{itemize}
            \item Initialise means and precision matrices of \(K\) Gaussian image-theme \(\{\bm{\mu}_{k}, \bm{\Lambda}_{k}\}_{k=1}^{K}\), where \(\bm{\mu}_{k} \in \mathbb{R}^{D}\), and  \(\bm{\Lambda}_{k} \in \mathbb{R}^{D \times D}\) is positive definite matrix
    		\item For task \(d\)-th in the collection of \(M\) tasks:
    		\begin{itemize}
    			\item Choose a task-theme mixture: \(\bm{\phi}_{d}~\sim~\mathrm{Dirichlet}_{L} \left( \bm{\phi}; \bm{\delta}\right) \)
    			\item For the \(c\)-th class in the \(d\)-th task:
    			\begin{itemize}
    				\item Choose a task-theme assignment: \(\mathbf{y}_{dc}~\sim~\mathrm{Categorical}(\mathbf{y}; \bm{\phi}_{d})\)
    				\item Choose an image-theme mixture: \(\bm{\theta}_{dc} \sim \mathrm{Dirichlet}_{K} \left( \bm{\theta}; \bm{\alpha}_{l} \right) \), where \({y_{dcl} = 1}\)
    				\item For image \(n\)-th in class \(c\)-th of task \(d\)-th:
    				\begin{itemize}
    					\item Choose an image-theme assignment: \(\mathbf{z}_{dcn}~\sim~\mathrm{Categorical} \left( \mathbf{z}; \bm{\theta}_{dc} \right)\)
    					\item Choose an image: \(\mathbf{x}_{dcn} \sim \mathcal{N}\left(\mathbf{x}; \bm{\mu}_{k}, \bm{\Lambda}_{k}^{-1}\right) \), where: \(z_{dcnk} = 1\).
    				\end{itemize}
    			\end{itemize}
    		\end{itemize}
    	\end{itemize}
    	
    	If the \(K\) Gaussian image-themes \( \{ (\bm{\mu}_k, \bm{\Lambda}_k) \} _{k=1}^K\), and the Dirichlet concentration \(\{ \bm{\alpha} \}_{l=1}^{L}\) for each class are known, we can infer the mixture parameter \(\bm{\phi}_{d}\) based on the observed images \(\mathbf{x}_{d}\) of any arbitrary task \(d\)-th to represent that task in the latent task-theme simplex. This representation enables the possibility of performing further analysis, such as measuring distances between tasks. Hence, our objective is to learn these parameters from the \(M\) given classification tasks. In short, our objective is to maximise log-likelihood:
    	\begin{equation}
            \max_{\bm{\mu}, \bm{\Sigma}, \bm{\alpha}} \ln p(\mathbf{x} | \bm{\mu}, \bm{\Sigma}, \bm{\alpha}).
            \label{eq:mle}
        \end{equation}
    	
    	Due to the complexity of the graphical model with latent variables as shown in \figureautorefname~\ref{fig:cclda}, the inference for the likelihood in \eqref{eq:mle} is intractable, and therefore, the estimation must rely on approximate inference. Current methods to approximate the posterior of LDA-based models fall into two main categories: sampling~\cite{griffiths2004finding,canini2009online} and optimisation~\cite{blei2003latent,teh2007collapsed}. Each approach has strengths and weaknesses, where the choice mostly depends on the application of interest. For the problem of task similarity where \(M\) is very large, the optimisation approach, and in particular, the mean-field VI, is preferable due to its efficiency and scalability to large data sets. In this paper, VI is used to infer the parameters of interest.
    	
    	The log-likelihood of interest can be lower-bounded by Jensen's inequality. The lower-bound is often known as evidence lower-bound (ELBO) and can be expressed as:
    	\begin{equation}
    	    \begin{aligned}[b]
    	     \mathsf{L} & = \mathbb{E}_{q} \left[ \ln p(\mathbf{x}, \bm{\phi}, \mathbf{y}, \bm{\theta}, \mathbf{z} | \bm{\delta}, \bm{\alpha}, \bm{\mu}, \bm{\Sigma}) \right] - \mathbb{E}_{q} \left[ q(\bm{\phi}, \mathbf{y}, \bm{\theta}, \mathbf{z}) \right].
    	     \end{aligned}
    	     \label{eq:elbo}
    	\end{equation}
    	
    	Following the conventional variational inference for LDA~\cite{blei2003latent}, we choose a fully factorised variational distribution \(q\) as our variational posterior:
    	\begin{equation}
    	    \begin{aligned}[b]
    		    q(\bm{\phi}, \mathbf{y}, \bm{\theta}, \mathbf{z}) & = \prod_{d=1}^{M} q(\bm{\phi}_{d}; \bm{\lambda}_{d}) \prod_{c=1}^{C} q(\mathbf{y}_{dc}; \bm{\eta}_{dc}) \, q(\bm{\theta}_{dc}; \bm{\gamma}_{dc}) \prod_{n=1}^{N} q(\mathbf{z}_{dcn}; \mathbf{r}_{dcn}),
            \end{aligned}
            \label{eq:q}
    	\end{equation}
    	where:
    	\begin{align*}
    		q(\bm{\phi}_{d}; \bm{\lambda}_{d}) = \mathrm{Dirichlet}_{L} \left(\bm{\phi}_{d}; \bm{\lambda}_{d} \right) & \qquad q(\mathbf{y}_{dc}; \bm{\eta}_{dc}) = \mathrm{Categorical}\left(\mathbf{y}_{dc}; \bm{\eta}_{dc}\right) \\
    		q(\bm{\theta}_{dc}; \bm{\gamma}_{dc}) = \mathrm{Dirichlet}_{K} \left( \bm{\theta}_{dc}; \bm{\gamma}_{dc} \right) & \qquad q(\mathbf{z}_{dcn}; \mathbf{r}_{dcn}) = \mathrm{Categorical} \left(\mathbf{z}_{dcn}; \mathbf{r}_{dcn} \right).
    	\end{align*}
    	
    	Given the variational distribution \(q\) defined in Eq.~\eqref{eq:q}, we can rewrite the ELBO as:
        \begin{equation}
            \begin{aligned}
                \mathsf{L} & = \mathbb{E}_{q} \left[ \ln p(\mathbf{x} | \mathbf{z}, \bm{\mu}, \bm{\Sigma}) + \ln p(\mathbf{z} | \bm{\theta}) + \ln p(\bm{\theta} | \mathbf{y}, \bm{\alpha}) + \textcolor{violet}{\ln p(\mathbf{y} | \bm{\phi})} + \textcolor{violet}{\ln p(\bm{\phi} | \bm{\delta})} \right.\\
                & \qquad \left. - \ln q(\mathbf{z}) - \ln q(\bm{\theta}) - \textcolor{violet}{\ln q(\mathbf{y})} - \textcolor{violet}{\ln q(\bm{\phi})} \right].
            \end{aligned}
            \label{eq:elbo_factorised}
        \end{equation}
    	Comparing to the conventional LDA~\cite[Eq.~(14)]{blei2003latent}, the ELBO in Eq.~\eqref{eq:elbo_factorised} contains 4 extra terms highlighted in \textcolor{violet}{violet}. The presence of those terms are due to the hierarchical structure of LDCC that takes the factor of classes (analogous to paragraphs) into the model.
    	
    	Instead of maximising likelihood, we maximise its lower-bound, resulting in an alternative objective function:
    	\begin{equation}
    	    \max_{\bm{\mu}, \bm{\Sigma}, \bm{\alpha}} \, \, \max_{\mathbf{r}, \bm{\gamma}, \bm{\eta}, \bm{\lambda}} \mathsf{L}.
    	\end{equation}
    	
    	Given the usage of prior conjugate, all of the terms in the ELBO can be evaluated straightforwardly (please refer to \appendixname{~\ref{apdx:elbo_derivation}}). The optimisation is based on gradient, and performed in two steps, resulting in a process analogous to the expectation-maximisation (EM) algorithm. In the E-step, the task-specific variational-parameters \(\mathbf{r}, \bm{\gamma}, \bm{\eta}\) and \(\bm{\lambda}\) are iteratively updated, while holding the meta-parameters \(\bm{\mu}, \bm{\Sigma}, \bm{\alpha}\) fixed. In the M-step, the meta-parameters are updated using the values of the task-specific variational-parameters obtained in the E-step. The inference for the meta image-themes are similar to the estimation of Gaussian mixture model~\cite[Chapter 9]{bishop2006pattern}. Please refer to \appendixname{~\ref{apdx:elbo_optimisation}} for more details.
    	
    	Conventionally, the iterative updates in the E-step and M-step require a full pass through the entire collection of tasks. This is, however, very slow and even infeasible since \(M\) is often in the magnitude of millions. We, therefore, propose an online VI inspired by the online learning for LDA~\cite{hoffman2010online} to infer the image-themes. When the \(d\)-th task  is observed, we perform EM to obtain the \say{task-specific} image-themes (denoted by a tilde on top of variables) that are locally optimal for that task. The \say{meta} image-themes of interest are then updated as a weighted average between their previous values and the \say{task-specific} values:
    	\begin{equation}
    	    \bm{\mu} \gets (1 - \rho_{d}) \bm{\mu} + \rho_{d} \Tilde{\bm{\mu}}, \quad \bm{\Sigma} \gets (1 - \rho_{d}) \bm{\Sigma} + \rho_{d} \Tilde{\bm{\Sigma}}, \quad \bm{\alpha} \gets \bm{\alpha} - \rho_{d} \mathbf{H}^{-1} \mathbf{g},
    	    \label{eq:online_update}
    	\end{equation}
    	where \(\rho_{d} = (\tau_{0} + d)^{-\tau_{1}}\) with \(\tau_{0} \ge 0\) and \(\tau_{1} \in (0.5, 1]\) \cite{hoffman2010online}, and \(\mathbf{g}\) is the gradient of \(\mathsf{L}\) w.r.t. \(\bm{\alpha}\), and \(\mathbf{H}\) is the Hessian matrix. Please refer to \appendixname{~\ref{apdx:learning_algorithm}} for the details of the online learning algorithm.
        
        Also, instead of updating the image-themes when observing a single task, we use multiple or a mini-batch of tasks to reduce noise. The mini-batch version requires a slight modification, where we calculate the average of all \say{task-specific} image-themes for the tasks in the same mini-batch, and use that as the \say{task-specific} value to update the corresponding \say{meta} image-theme.
        
        Given the image-themes \(\{\bm{\mu}_{k}, \bm{\Sigma}_{k} \}_{k=1}^{K} \) and the Dirichlet parameter \(\{ \bm{\alpha}_{l} \}_{l=1}^{L}\), we can represent a task by its variational Dirichlet posterior of the task-topic mixing coefficients \(q(\bm{\phi}_{d}; \bm{\lambda}_{d})\) in the latent task-theme simplex. This new representation of classification tasks has two advantages comparing to the recently proposed task representation Task2Vec~\cite{achille2019task2vec}: (i) it does not need any pre-trained networks, and (ii) the use of probability distribution, instead of a single value vector as in Task2Vec, allowing to include modelling uncertainty when representing tasks. In addition, we can utilise this representation to quantitatively analyse the similarity between two tasks through a divergence between \(q(\bm{\phi}_{d}; \bm{\lambda}_{d})\). 
        Commonly, symmetric distances, such as Jensen-Shannon divergence, Hellinger distance, or earth's mover distance are employed to calculate the divergence between distributions. However, it is argued that similarity should be represented as an asymmetric measure~\cite{tversky1977features}. This is reasonable in the context of transfer learning, since knowledge gained from learning a difficult task might significantly facilitate the learning of an easy task, but the reverse might not always have the same level of effectiveness. In light of asymmetric distance, we decide to use Kullback-Leibler (KL) divergence, denoted as \(D_{\mathrm{KL}}[. \Vert .]\). As \(D_{\mathrm{KL}} \left[ P \Vert Q \right]\) is defined as the information lost when using a code optimised for \(Q\) to encode the samples of \(P\), we, therefore, calculate \(D_{\mathrm{KL}} \left[ q(\bm{\phi}_{d}; \bm{\lambda}_{M + 1}) \Vert  q(\bm{\phi}_{d}; \bm{\lambda}_{d}) \right]\), where \(d \in \{1, \ldots, M\}\), to assess how the \(d\)-th training task differs from the learning of the novel \((M + 1)\)-th task.
    
        \paragraph{Correlation Diagram}
        We define a correlation diagram as a qualitative measure that represents visually the \say{performance effectiveness} for meta-learning algorithms. The diagram plots the expected classification accuracy as a function of KL divergence between testing and training tasks. Intuitively, the closer a testing task is from the training tasks, the higher the performance. Hence, we can use our proposed correlation diagram to qualitatively compare different meta-learning methods.
        
        A correlation diagram can be constructed by first calculating the average distance between each testing task, denoted as \(M + i\) subscript with \(i \in \mathbb{N}\), to all training tasks:
        \begin{equation*}
            \overline{D}_{M + i} = \frac{1}{M} \sum_{d=1}^{M} D_{\mathrm{KL}} \left[q(\bm{\phi}; \bm{\lambda}_{M + i}) \Vert q(\bm{\phi}; \bm{\lambda}_{d}) \right].
        \end{equation*}
        The obtained average distances are then grouped into \(J\) interval bins, each of size \(\triangle_{J}~=~\max_{i} \overline{D}_{M + i} /J\). Let \(B_{j}\) with \(j \in \{1, \ldots, J\}\) be the set of testing tasks that have their average KL distances falling within the interval \(I_{j} = \left((j - 1) \triangle_{J}, j \triangle_{J} \right]\). The distance of bin \(B_{j}\) is defined as:
        \begin{equation*}
            d(B_{j}) = \frac{1}{|B_{j}|} \sum_{i \in B_{j}} \overline{D}_{M + i}.
        \end{equation*}
        Next, a model trained on the training tasks is employed to evaluate the prediction accuracy \(a^{(v)}_{i}\) on all the testing tasks to obtain the accuracy for bin \( B_{j} \):
        \begin{equation*}
            a(B_{j}) = \frac{1}{|B_{j}|} \sum_{i \in B_{j}} a^{(v)}_{i}.
        \end{equation*}
        Finally, plotting \(d(B_{j})\) against \(a(B_{j})\) gives the desired correlation diagram (e.g., \figureautorefname{~\ref{fig:correlation_diagram}}).
        
    \section{Experiments}
\label{sec:experiment}
    We carry out two experiments -- correlation diagram and task selection -- to demonstrate the capability of the proposed approach. We evaluate the proposed approach on \(n\)-way classification tasks formed from two separated data sets: Omniglot~\cite{lake2015human} and mini-ImageNet~\cite{vinyals2016matching}. In this setting, a testing task is represented by a \(k\)-shot labelled data without the availability of unlabelled data following the transductive learning setting~\cite{dhillon2019baseline}. We evaluate the performance on several meta-learning algorithms, such as MAML~\cite{finn2017model}, Prototypical Networks~\cite{snell2017prototypical}, Amortised Meta-learner (ABML)~\cite{ravi2018amortized}, BMAML~\cite{yoon2018bayesian} and VAMPIRE~\cite{nguyen2020uncertainty}, to verify the distance-performance correlation using our proposed method.
    
    For Omniglot, we follow the pre-processing steps as in few-shot image classification without any data augmentation, and use the standard train-test split in the original paper to prevent information leakage. For mini-ImageNet, we follow the common train-test split with 80 classes for training and 20 classes for testing~\cite{ravi2017optimization}. Since the dimension of raw images in mini-ImageNet is large, we employ the 640-dimensional features extracted from a wide-residual-network~\cite{rusu2019meta} to ease the calculation.
    
    We follow Algorithm~\ref{alg:online_ldcc} in \appendixname{~\ref{apdx:learning_algorithm}} to obtain the posterior of the image-theme using tasks in training set. We use \(L = 4\) task-themes and \(K = 8\) image-themes for both data sets. The Dirichlet distribution for task-theme mixture, \(\mathrm{Dirchlet}_{L}(\bm{\phi}_{d} | \bm{\delta}\), is chosen to be symmetric with \(\delta = 0.5\). The parameter inference, or training, is carried out with 16 images per class while varying the number of classes between 5 to 10 to fit into the memory of a Nvidia 1080 Ti GPU. The inference of the variational parameter \(\bm{\lambda}\) is done on all available labelled data in each class (\(20\) for Omniglot and \(600\) for mini-ImageNet). Note that this is used for the correlation diagram demonstration. For the task selection, this number matches the number of shots in the few-shot learning setting\footnote{Implementation can be found at \url{https://github.com/cnguyen10/similarity\_classification\_tasks}}.
    
    For the evaluation on meta-learning algorithms, we use a similar 4 convolutional module network to train on Omniglot~\cite{vinyals2016matching,finn2017model,snell2017prototypical}, while using a fully connected network with 1 hidden layer consisting of 128 units to train on the extracted features of mini-ImageNet~\cite{nguyen2020uncertainty}.
    
    Note that the numbers of tasks formed from the two data sets are very large. For Omniglot, approximate \(6.8 \times 10^{12}\) and \(10^{12}\) unique tasks can be generated from the training and testing sets, respectively. For mini-ImageNet, these numbers are slightly more manageable with about \(2.4 \times 10^{6}\) unique tasks for training, and \(15,504\) tasks for testing. To reduce the computation and facilitate the analysis, we randomly select 1 million Omniglot tasks for training, and \(20,000\) tasks for testing. For mini-ImageNet, we select 2 million tasks for training and \(15,504\) tasks for testing.
    
    \subsection{Correlation Diagram}
    
    \begin{figure*}[t]
        \centering
        \begin{subfigure}{0.45 \textwidth}
            \centering
            \resizebox{\textwidth}{!}{
                \begin{tikzpicture}[]

\definecolor{color0}{rgb}{0.12156862745098,0.466666666666667,0.705882352941177}
\definecolor{color1}{rgb}{1,0.498039215686275,0.0549019607843137}
\definecolor{color2}{rgb}{0.172549019607843,0.627450980392157,0.172549019607843}
\definecolor{color3}{rgb}{0.83921568627451,0.152941176470588,0.156862745098039}
\definecolor{color4}{rgb}{0.580392156862745,0.403921568627451,0.741176470588235}

\begin{axis}[
legend cell align={left},
legend style={fill opacity=0.8, draw opacity=1, text opacity=1, draw=none},
tick align=outside,
tick pos=both,
x grid style={white!69.01960784313725!black},
xlabel={KL divergence from test to train},
xmin=0.8018515794, xmax=5.1246649886,
xtick style={color=black},
y grid style={white!69.01960784313725!black},
ylabel={Prediction accuracy},
ymin=0.95514979895, ymax=0.98328278405,
ytick style={color=black},
ytick={0.955,0.96,0.965,0.97,0.975,0.98,0.985},
yticklabels={0.955,0.960,0.965,0.970,0.975,0.980,0.985}
]
\addplot [semithick, color0, mark=*, mark size=3, mark options={solid}]
table {%
0.998343098 0.970667112
1.31499724 0.970730504
1.73650169 0.969635752
2.18051967 0.968153083
2.62905746 0.965035781
3.08299814 0.966122533
3.53206666 0.961474247
3.97433777 0.965170654
4.43768848 0.961515038
4.92817347 0.958165414
};
\addlegendentry{ProtoNet}
\addplot [semithick, color1, mark=triangle*, mark size=3, mark options={solid,rotate=180}]
table {%
0.998343098 0.975021492
1.31499724 0.972281868
1.73650169 0.971431216
2.18051967 0.970176447
2.62905746 0.96856952
3.08299814 0.966957237
3.53206666 0.964987335
3.97433777 0.965653907
4.43768848 0.959684211
4.92817347 0.956428571
};
\addlegendentry{MAML}
\addplot [semithick, color2, mark=triangle*, mark size=3, mark options={solid}]
table {%
0.998343098 0.980144781
1.31499724 0.9751952
1.73650169 0.974905812
2.18051967 0.972461487
2.62905746 0.966740034
3.08299814 0.965666118
3.53206666 0.964005629
3.97433777 0.966188198
4.43768848 0.962669173
4.92817347 0.958571429
};
\addlegendentry{ABML}
\addplot [semithick, color3, mark=triangle*, mark size=3, mark options={solid,rotate=270}]
table {%
0.998343098 0.980683637
1.31499724 0.979006757
1.73650169 0.977964153
2.18051967 0.973143519
2.62905746 0.969689333
3.08299814 0.969877467
3.53206666 0.967799043
3.97433777 0.969208931
4.43768848 0.967631579
4.92817347 0.965894737
};
\addlegendentry{BMAML}
\addplot [semithick, color4, mark=square*, mark size=3, mark options={solid}]
table {%
0.998343098 0.982004012
1.31499724 0.97901661
1.73650169 0.972784022
2.18051967 0.971461487
2.62905746 0.968740034
3.08299814 0.967666118
3.53206666 0.970005629
3.97433777 0.964188198
4.43768848 0.960669173
4.92817347 0.960571429
};
\addlegendentry{VAMPIRE}
\end{axis}

                \end{tikzpicture}
            }
            \caption{Omniglot}
            \label{fig:correlation_omniglot}
        \end{subfigure}
        \hfill
        \begin{subfigure}{0.45 \textwidth}
            \centering
            \resizebox{\textwidth}{!}{
                \begin{tikzpicture}

\definecolor{color0}{rgb}{0.12156862745098,0.466666666666667,0.705882352941177}
\definecolor{color1}{rgb}{1,0.498039215686275,0.0549019607843137}
\definecolor{color2}{rgb}{0.172549019607843,0.627450980392157,0.172549019607843}
\definecolor{color3}{rgb}{0.83921568627451,0.152941176470588,0.156862745098039}
\definecolor{color4}{rgb}{0.580392156862745,0.403921568627451,0.741176470588235}

\begin{axis}[
tick align=outside,
tick pos=both,
x grid style={white!69.01960784313725!black},
xlabel={KL divergence from test to train},
xmin=0.180070755, xmax=2.498431745,
xtick style={color=black},
y grid style={white!69.01960784313725!black},
ylabel={Prediction accuracy},
ymin=0.5102780362, ymax=0.6345668118,
ytick style={color=black}
]
\addplot [semithick, color0, mark=*, mark size=3, mark options={solid}]
table {%
0.2854508 0.593095636
0.46133616 0.588150791
0.710168811 0.582093974
0.955166249 0.578580677
1.20270308 0.578212203
1.44066921 0.571644834
1.68019509 0.563948963
1.93140635 0.553984418
2.18265238 0.541368153
2.3930517 0.536019837
};
\addplot [semithick, color1, mark=triangle*, mark size=3, mark options={solid,rotate=180}]
table {%
0.2854508 0.595210093
0.46133616 0.592075198
0.710168811 0.582229685
0.955166249 0.575907228
1.20270308 0.569081117
1.44066921 0.57407833
1.68019509 0.553331266
1.93140635 0.555192782
2.18265238 0.533052707
2.3930517 0.515927526
};
\addplot [semithick, color2, mark=triangle*, mark size=3, mark options={solid}]
table {%
0.2854508 0.612406892
0.46133616 0.605769265
0.710168811 0.601543616
0.955166249 0.596806021
1.20270308 0.593803371
1.44066921 0.584549188
1.68019509 0.583350346
1.93140635 0.570208019
2.18265238 0.565178472
2.3930517 0.55740548
};
\addplot [semithick, color3, mark=triangle*, mark size=3, mark options={solid,rotate=270}]
table {%
0.2854508 0.628917322
0.46133616 0.620390874
0.710168811 0.616816174
0.955166249 0.606001276
1.20270308 0.60088618
1.44066921 0.594176425
1.68019509 0.589624612
1.93140635 0.578779447
2.18265238 0.58722474
2.3930517 0.561425906
};
\addplot [semithick, color4, mark=square*, mark size=3, mark options={solid}]
table {%
0.2854508 0.601331423
0.46133616 0.599500218
0.710168811 0.599772962
0.955166249 0.592460733
1.20270308 0.592100457
1.44066921 0.593868613
1.68019509 0.586704762
1.93140635 0.594603175
2.18265238 0.56
2.3930517 0.555686275
};
\end{axis}

                \end{tikzpicture}
            }
            \caption{mini-ImageNet}
            \label{fig:correlation_miniImageNet}
        \end{subfigure}
        \caption{Correlation diagram plots the average accuracy predicted by meta-learning algorithms as a function of the average KL divergence of each task in the testing set to all tasks in the training set on the 5-way 1-shot setting.}
        \label{fig:correlation_diagram}
    \end{figure*}
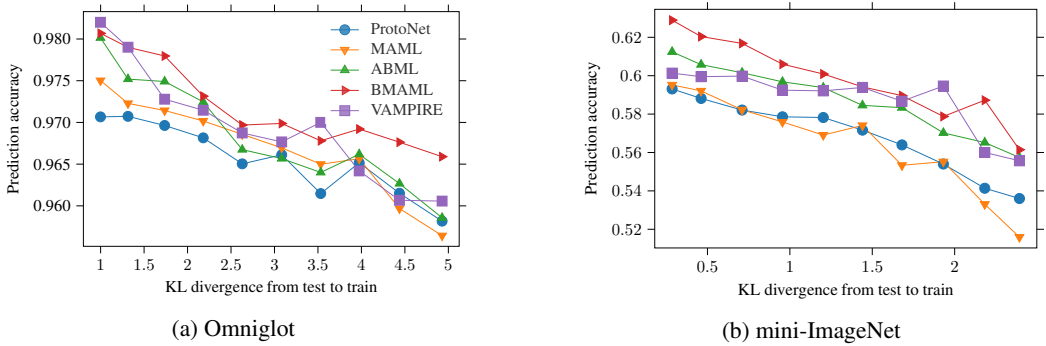
    To construct the correlation diagram, we train a continuous LDCC on \(n\)-way 16-shot setting (\(n\) varies from 5 to 10), and then infer the variational parameter \(\bm{\lambda}\) of the task-theme mixture \(\bm{\phi}\). The inferred \(\lambda\) is used to calculate the KL divergence distance between testing tasks to all training tasks. Note that the continuous LDCC is only trained on the training tasks. We then separately evaluate the performance of different meta-learning algorithms on the same 5-way 1-shot setting, and plot the correlation diagram in \figureautorefname~\ref{fig:correlation_diagram}. The results of the performance versus the task distance (or similarity) agree well with the common intuition: the testing tasks closer to the training tasks have higher prediction accuracy. Note that this observation is consistent across several meta-learning methods. It is also interesting to notice that some methods are more robust than others with respect to the dissimilarity between training and testing tasks.
    
    \subsection{Task Selection}
    
        \begin{figure}[t]
            \centering
            \resizebox{0.45\textwidth}{!}{
                \begin{tikzpicture}

\definecolor{color0}{rgb}{0.172549019607843,0.627450980392157,0.172549019607843}
\definecolor{color1}{rgb}{0.12156862745098,0.466666666666667,0.705882352941177}
\definecolor{color2}{rgb}{1,0.498039215686275,0.0549019607843137}
\definecolor{color3}{rgb}{0.737254901960784,0.741176470588235,0.133333333333333}

\begin{axis}[
legend cell align={left},
legend style={fill opacity=0.8, draw opacity=1, text opacity=1, at={(0.03,0.97)}, anchor=north west, draw=none},
tick align=outside,
tick pos=left,
x grid style={white!69.0196078431373!black},
xmin=-0.62625, xmax=4.62625,
xtick style={color=black},
xtick={0,1,2,3,4},
xticklabels={ProtoNet,MAML,ABML,BMAML,VAMPIRE},
y grid style={white!69.0196078431373!black},
ylabel={Prediction accuracy (\%)},
ymin=58, ymax=63,
ytick style={color=black}
]
\draw[draw=none,fill=color1,fill opacity=1.] (axis cs:-0.3875,0) rectangle (axis cs:-0.1375,60.29);
\addlegendimage{ybar,ybar legend,draw=none,fill=color1,fill opacity=1.};
\addlegendentry{Selective}

\draw[draw=none,fill=color1,fill opacity=1.] (axis cs:0.6125,0) rectangle (axis cs:0.8625,61.09);
\draw[draw=none,fill=color1,fill opacity=1.] (axis cs:1.6125,0) rectangle (axis cs:1.8625,61.82);
\draw[draw=none,fill=color1,fill opacity=1.] (axis cs:2.6125,0) rectangle (axis cs:2.8625,62.52);
\draw[draw=none,fill=color1,fill opacity=1.] (axis cs:3.6125,0) rectangle (axis cs:3.8625,61.47);
\draw[draw=none,fill=color2,fill opacity=1.] (axis cs:-0.125,0) rectangle (axis cs:0.125,60.01);
\addlegendimage{ybar,ybar legend,draw=none,fill=color2,fill opacity=1.};
\addlegendentry{Task2Vec}

\draw[draw=none,fill=color2,fill opacity=1.] (axis cs:0.875,0) rectangle (axis cs:1.125,61.17);
\draw[draw=none,fill=color2,fill opacity=1.] (axis cs:1.875,0) rectangle (axis cs:2.125,61.23);
\draw[draw=none,fill=color2,fill opacity=1.] (axis cs:2.875,0) rectangle (axis cs:3.125,61.98);
\draw[draw=none,fill=color2,fill opacity=1.] (axis cs:3.875,0) rectangle (axis cs:4.125,61.05);
\draw[draw=none,fill=color3,fill opacity=1.] (axis cs:0.1375,0) rectangle (axis cs:0.3875,58.58);
\addlegendimage{ybar,ybar legend,draw=none,fill=color3,fill opacity=1.};
\addlegendentry{Random}

\draw[draw=none,fill=color3,fill opacity=1.] (axis cs:1.1375,0) rectangle (axis cs:1.3875,59.4);
\draw[draw=none,fill=color3,fill opacity=1.] (axis cs:2.1375,0) rectangle (axis cs:2.3875,59.75);
\draw[draw=none,fill=color3,fill opacity=1.] (axis cs:3.1375,0) rectangle (axis cs:3.3875,61.21);
\draw[draw=none,fill=color3,fill opacity=1.] (axis cs:4.1375,0) rectangle (axis cs:4.3875,59.62);
\path [draw=color0, semithick]
(axis cs:0.2625,57.3298917794047)
--(axis cs:0.2625,59.8301082205953);

\path [draw=color0, semithick]
(axis cs:1.2625,58.1110857592532)
--(axis cs:1.2625,60.6889142407468);

\path [draw=color0, semithick]
(axis cs:2.2625,58.6883781629978)
--(axis cs:2.2625,60.8116218370022);

\path [draw=color0, semithick]
(axis cs:3.2625,60.1095721428462)
--(axis cs:3.2625,62.3104278571538);

\path [draw=color0, semithick]
(axis cs:4.2625,58.5417470115043)
--(axis cs:4.2625,60.6982529884957);

\addplot [semithick, color0, mark=-, mark size=5, mark options={solid}, only marks]
table {%
0.2625 57.3298917794047
1.2625 58.1110857592532
2.2625 58.6883781629978
3.2625 60.1095721428462
4.2625 58.5417470115043
};
\addplot [semithick, color0, mark=-, mark size=5, mark options={solid}, only marks]
table {%
0.2625 59.8301082205953
1.2625 60.6889142407468
2.2625 60.8116218370022
3.2625 62.3104278571538
4.2625 60.6982529884957
};
\end{axis}

                \end{tikzpicture}
            }
            \caption{The prediction accuracy of several meta-learning methods on 5-way 5-shot mini-ImageNet testing tasks when training tasks are pro-actively selected outperforms the un-selective approaches, and slightly better than Task2Vec. The error bars on the un-selective cases represent the 95\% confident intervals calculated on the 50 trials of random task selection.}
            \label{fig:task_selection}
        \end{figure}
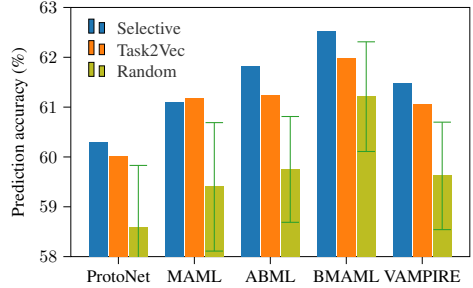
    
        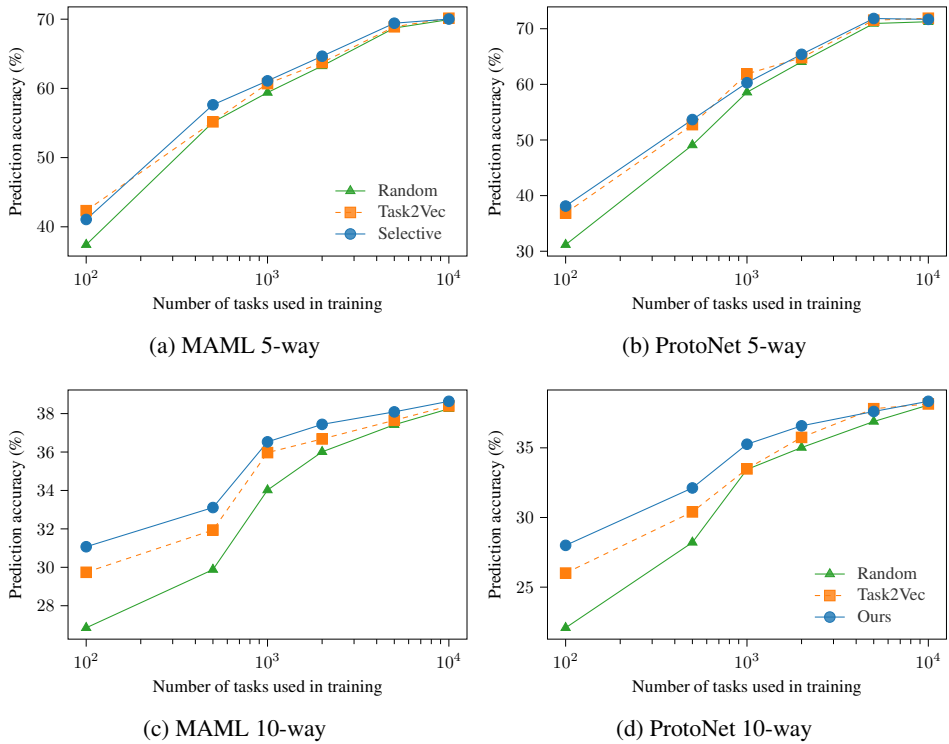
\begin{figure*}[t]
            \centering
            \begin{subfigure}{0.45 \textwidth}
                \centering
                \resizebox{\textwidth}{!}{
                    \begin{tikzpicture}

\definecolor{color0}{rgb}{0.12156862745098,0.466666666666667,0.705882352941177}
\definecolor{color1}{rgb}{0.172549019607843,0.627450980392157,0.172549019607843}
\definecolor{color2}{rgb}{1,0.498039215686275,0.0549019607843137}

\begin{axis}[
legend cell align={left},
legend style={fill opacity=0.8, draw opacity=1, text opacity=1, at={(0.97,0.03)}, anchor=south east, draw=none},
log basis x={10},
tick align=outside,
tick pos=left,
x grid style={white!69.0196078431373!black},
xlabel={Number of tasks used in training},
xmin=79.4328234724281, xmax=12589.2541179417,
xmode=log,
xtick style={color=black},
xtick={1,10,100,1000,10000,100000,1000000},
xticklabels={\(\displaystyle {10^{0}}\),\(\displaystyle {10^{1}}\),\(\displaystyle {10^{2}}\),\(\displaystyle {10^{3}}\),\(\displaystyle {10^{4}}\),\(\displaystyle {10^{5}}\),\(\displaystyle {10^{6}}\)},
minor xtick={200, 300,..., 900, 2000, 3000, ..., 9000},
y grid style={white!69.0196078431373!black},
ylabel={Prediction accuracy (\%)},
ymin=35.7735, ymax=71.7765,
ytick style={color=black}
]
\addplot [semithick, color1, mark=triangle*, mark size=3, mark options={solid}]
table {%
100 37.41
500 55.12
1000 59.4
2000 63.24
5000 68.72
10000 69.94
};
\addlegendentry{Random}
\addplot [semithick, color2, dashed, mark=square*, mark size=3, mark options={solid}]
table {%
100 42.33
500 55.17
1000 60.69
2000 63.73
5000 68.9
10000 70.14
};
\addlegendentry{Task2Vec}
\addplot [semithick, color0, mark=*, mark size=3, mark options={solid}]
table {%
100 41.06
500 57.63
1000 61.09
2000 64.65
5000 69.42
10000 70.02
};
\addlegendentry{Selective}
\end{axis}

                    \end{tikzpicture}
                }
                \caption{MAML 5-way}
                \label{fig:maml_5way}
            \end{subfigure}
            \begin{subfigure}{0.45 \textwidth}
                \centering
                \resizebox{\textwidth}{!}{
                    \begin{tikzpicture}

\definecolor{color0}{rgb}{0.12156862745098,0.466666666666667,0.705882352941177}
\definecolor{color1}{rgb}{0.172549019607843,0.627450980392157,0.172549019607843}
\definecolor{color2}{rgb}{1,0.498039215686275,0.0549019607843137}
\begin{axis}[
log basis x={10},
tick align=outside,
tick pos=left,
x grid style={white!69.0196078431373!black},
xlabel={Number of tasks used in training},
xmin=79.4328234724281, xmax=12589.2541179417,
xmode=log,
xtick style={color=black},
xtick={1,10,100,1000,10000,100000,1000000},
xticklabels={\(\displaystyle {10^{0}}\),\(\displaystyle {10^{1}}\),\(\displaystyle {10^{2}}\),\(\displaystyle {10^{3}}\),\(\displaystyle {10^{4}}\),\(\displaystyle {10^{5}}\),\(\displaystyle {10^{6}}\)},
minor xtick={200, 300,..., 900, 2000, 3000, ..., 9000},
y grid style={white!69.0196078431373!black},
ylabel={Prediction accuracy (\%)},
ymin=29.134, ymax=73.926,
ytick style={color=black}
]
\addplot [semithick, color1, mark=triangle*, mark size=3, mark options={solid}]
table {%
100 31.17
500 49.09
1000 58.58
2000 64.05
5000 70.94
10000 71.26
};
\addplot [semithick, color2, dashed, mark=square*, mark size=3, mark options={solid}]
table {%
100 36.86
500 52.76
1000 61.87
2000 64.79
5000 71.54
10000 71.89
};
\addplot [semithick, color0, mark=*, mark size=3, mark options={solid}]
table {%
100 38.12
500 53.67
1000 60.29
2000 65.4
5000 71.84
10000 71.68
};
\end{axis}

                    \end{tikzpicture}
                }
                \caption{ProtoNet 5-way}
                \label{fig:protonet_5way}
            \end{subfigure}
            \hfill\\
            \vspace{1em}
            \begin{subfigure}{0.45 \textwidth}
                \centering
                \resizebox{\textwidth}{!}{
                    \begin{tikzpicture}

\definecolor{color0}{rgb}{0.12156862745098,0.466666666666667,0.705882352941177}
\definecolor{color1}{rgb}{0.172549019607843,0.627450980392157,0.172549019607843}
\definecolor{color2}{rgb}{1,0.498039215686275,0.0549019607843137}

\begin{axis}[
log basis x={10},
tick align=outside,
tick pos=left,
x grid style={white!69.0196078431373!black},
xlabel={Number of tasks used in training},
xmin=79.4328234724281, xmax=12589.2541179417,
xmode=log,
xtick style={color=black},
xtick={1,10,100,1000,10000,100000,1000000},
xticklabels={\(\displaystyle {10^{0}}\),\(\displaystyle {10^{1}}\),\(\displaystyle {10^{2}}\),\(\displaystyle {10^{3}}\),\(\displaystyle {10^{4}}\),\(\displaystyle {10^{5}}\),\(\displaystyle {10^{6}}\)},
minor xtick={200, 300,..., 900, 2000, 3000, ..., 9000},
y grid style={white!69.0196078431373!black},
ylabel={Prediction accuracy (\%)},
ymin=26.2605, ymax=39.2295,
ytick style={color=black}
]
\addplot [semithick, color1, mark=triangle*, mark size=3, mark options={solid}]
table {%
100 26.85
500 29.88
1000 34.02
2000 36.01
5000 37.42
10000 38.25
};
\addplot [semithick, color2, dashed, mark=square*, mark size=3, mark options={solid}]
table {%
100 29.74
500 31.94
1000 35.97
2000 36.68
5000 37.65
10000 38.4
};
\addplot [semithick, color0, mark=*, mark size=3, mark options={solid}]
table {%
100 31.07
500 33.11
1000 36.53
2000 37.44
5000 38.09
10000 38.64
};
\end{axis}

                    \end{tikzpicture}
                }
                \caption{MAML 10-way}
                \label{fig:maml_10way}
            \end{subfigure}
            \begin{subfigure}{0.45 \textwidth}
                \centering
                \resizebox{\textwidth}{!}{
                    \begin{tikzpicture}

\definecolor{color0}{rgb}{0.12156862745098,0.466666666666667,0.705882352941177}
\definecolor{color1}{rgb}{0.172549019607843,0.627450980392157,0.172549019607843}
\definecolor{color2}{rgb}{1,0.498039215686275,0.0549019607843137}

\begin{axis}[
legend cell align={left},
legend style={fill opacity=0.8, draw opacity=1, text opacity=1, at={(0.97,0.03)}, anchor=south east, draw=none},
log basis x={10},
tick align=outside,
tick pos=left,
x grid style={white!69.0196078431373!black},
xlabel={Number of tasks used in training},
xmin=79.4328234724281, xmax=12589.2541179417,
xmode=log,
xtick style={color=black},
xtick={1,10,100,1000,10000,100000,1000000},
xticklabels={\(\displaystyle {10^{0}}\),\(\displaystyle {10^{1}}\),\(\displaystyle {10^{2}}\),\(\displaystyle {10^{3}}\),\(\displaystyle {10^{4}}\),\(\displaystyle {10^{5}}\),\(\displaystyle {10^{6}}\)},
minor xtick={200, 300,..., 900, 2000, 3000, ..., 9000},
y grid style={white!69.0196078431373!black},
ylabel={Prediction accuracy (\%)},
ymin=21.267, ymax=39.153,
ytick style={color=black}
]
\addplot [semithick, color1, mark=triangle*, mark size=3, mark options={solid}]
table {%
100 22.08
500 28.21
1000 33.45
2000 35.02
5000 36.89
10000 38.07
};
\addlegendentry{Random};
\addplot [semithick, color2, dashed, mark=square*, mark size=3, mark options={solid}]
table {%
100 26.01
500 30.4
1000 33.49
2000 35.75
5000 37.81
10000 38.15
};
\addlegendentry{Task2Vec}
\addplot [semithick, color0, mark=*, mark size=3, mark options={solid}]
table {%
100 28
500 32.12
1000 35.26
2000 36.58
5000 37.61
10000 38.34
};
\addlegendentry{Ours}
\end{axis}

                    \end{tikzpicture}
                }
                \caption{ProtoNet 10-way}
                \label{fig:protonet_10way}
            \end{subfigure}
            \caption{The proposed task-selective approach outperforms the randomly chosen training tasks, and shows slightly better results than Task2Vec when varying the number of classes within a classification task as well as the number of training tasks.}
            \label{fig:varying_num_tasks_ways}
        \end{figure*}
        
        We show that when there is a constraint on the number of training tasks, selecting tasks based on the proposed similarity outperforms the un-selective one that randomly selects training tasks. To demonstrate, we assume that one can pick a small number of mini-ImageNet tasks from the whole training set to train a meta-learning model, and evaluate on all tasks in the testing set. In the selective case, we use the LDCC model trained on all training tasks to infer the variational mixture parameters \(\bm{\lambda}\) for all training and testing tasks. We then pick the training tasks that are closest to all the testing tasks using the proposed KL divergence, and use them to train a meta-learning model. In the un-selective case, we randomly select the same number of training tasks without measuring any similarity. We also include Task2Vec as a baseline for the selective case to compare with our proposed approach. As the experiment is based on extracted features of mini-ImageNet, it is difficult to adapt to some common pre-trained networks, which is used as a \say{probe} network in Task2Vec. To work around, we use MAML to train a fully-connected network with three hidden layers on the training set under 5-way 5-shot setting, and use the feature extractor (excluding the last layer) of this network as the \say{probe} network for Task2Vec. This modelling approach results in a 3-D Task2Vec representation which is the same dimension as \(\bm{\phi}_{d}\) in the continuous LDCC, and hence, can be compared fairly. In addition, we directly calculate the diagonal of Fisher information matrix of the \say{probe} network without using the proposed approximation in Task2Vec to reduce the complexity of hyper-parameter tuning.
        
        \figureautorefname~\ref{fig:task_selection} shows the accuracy results tested on \(15,504\) mini-ImageNet testing tasks on the 5-way 5-shot setting for models trained on 1,000 training tasks. We also report the 95\% confident interval for the case of random task selection. Statistically, meta-learning methods trained on tasks selected from our proposed solution outperform the un-selective cases, and slightly better than Task2Vec, especially for the probabilistic meta-learning methods such as BMAML, ABML and VAMPIRE.
        
        To study the effects induced by the number of training tasks, and the number of ways within each task, we run an extensive experiment with a similar 5-shot setting, but varying the number of training tasks and ways, and plot the results in \figureautorefname~\ref{fig:varying_num_tasks_ways}. In general, the proposed approach out-performs the un-selective approach, and is  slightly better than Task2Vec.
        
        Despite promising results, there are some limitations of our proposed task selection. The proposed approach requires a sufficient number of labelled data in the testing tasks. More specifically, we need 5 labelled images per class, so that the trained LDCC model can correctly infer \(\bm{\lambda}\). Further reduction in the number of labelled data in the test set might result in a poor estimation of \(\bm{\lambda}\), hindering the task selection process. This is a well-known issue in LDA and its variations, which do not work well for short texts. Nevertheless, the assumption of 5-shot setting, which shows a promising result for task selection, is still reasonable in many few-shot learning applications.
    \section{Conclusion}
\label{sec:conclusion}
    We propose a generative approach based on the continuous LDCC adopted in topic modelling to model classification tasks. Under this modelling approach, a classification task can be expressed as a finite mixture model of Gaussian distributions, whose components are shared across all tasks. This new representation of classification tasks allows one to quantify the similarity between tasks through the asymmetric KL divergence. We also introduce a task selection strategy based on the proposed task similarity, and demonstrate its superiority in meta-learning comparing to the conventional approach where training tasks are randomly selected.
    
    \section*{Acknowledgement}
        This work was supported with supercomputing resources provided by the Phoenix HPC service at the University of Adelaide.
    
    \section*{Broader impact}
        The proposed approach is helpful in transfer-learning tasks, especially when the amount of training data for the testing task is limited. By representing tasks in the topic space, the proposed approach allows to assess task similarity and provide insightful understanding when transfer-learning will be effective. This has the benefit of saving costs on data collection and annotation for the testing task. However, the trade-off is related to the computational cost involved in the training of the LDCC model for computing the task-to-task similarities.
    
    \printbibliography
    
    \newpage
    \onecolumn
\appendix
    
    
\section{Calculation of each term in the lower-bound}
\label{apdx:elbo_derivation}
    This section presents the calculation of for each term of the ELBO in \eqref{eq:elbo_factorised}. Note that the variational distribution \(q\) is defined in \eqref{eq:q}.
    
    \subsection{\texorpdfstring{\(\mathbb{E}_{q} \left[ \ln p(\mathbf{x} | \mathbf{z}, \bm{\mu}, \bm{\Sigma}) \right]\)}{}}
        \begin{equation}
            \begin{aligned}
                & \mathbb{E}_{q} \left[ \ln p(\mathbf{x} | \mathbf{z}, \bm{\mu}, \bm{\Sigma}) \right] \\
                & = \sum_{d=1}^{M} \sum_{\mathbf{z}} q(\mathbf{z}; \mathbf{r}) \ln \ln p(\mathbf{x} | \mathbf{z}, \bm{\mu}, \bm{\Sigma}) \\
                & = \sum_{d=1}^{M} \sum_{c=1}^{C} \sum_{n=1}^{N} \sum_{k=1}^{K} q(z_{dcnk} = 1; r_{dcnk}) \ln p(\mathbf{x}_{dcn} | \bm{\mu}_{k}, \bm{\Sigma}_{k}) \\
                & = - \frac{1}{2}\sum_{d=1}^{M} \sum_{c=1}^{C} \sum_{n=1}^{N} \sum_{k=1}^{K} r_{dcnk} \left[ D \ln(2\pi) + \ln |\bm{\Sigma}_{k}| + (\mathbf{x}_{dcn} - \bm{\mu}_{k})^{\top} \bm{\Sigma}_{k}^{-1} (\mathbf{x}_{dcn} - \bm{\mu}_{k})  \right].
            \end{aligned}
        \end{equation}
        
    \subsection{\texorpdfstring{\(\mathbb{E}_{q} \left[ \ln p(\mathbf{z} | \bm{\theta}) \right]\)}{}}
        \begin{equation}
            \begin{aligned}
                \mathbb{E}_{q} \left[ \ln p(\mathbf{z} | \bm{\theta}) \right] & = \sum_{d=1}^{M} \sum_{c=1}^{C} \sum_{n=1}^{N} \sum_{k=1}^{K} q(z_{dcnk} = 1; r_{dcnk}) \int q(\bm{\theta}_{dc}; \bm{\gamma}_{dc}) \ln p(z_{dcnk} = 1 | \bm{\theta}_{dc}) \dd{\bm{\theta}_{dc}} \\
                & = \sum_{d=1}^{M} \sum_{c=1}^{C} \sum_{n=1}^{N} \sum_{k=1}^{K} r_{dcnk} \int \mathrm{Dirichlet}(\bm{\theta}_{dc}; \bm{\gamma}_{dc}) \ln \theta_{dck} \dd{\theta_{dck}} \\
                & = \sum_{d=1}^{M} \sum_{c=1}^{C} \sum_{n=1}^{N} \sum_{k=1}^{K} r_{dcnk} \ln \Tilde{\theta}_{dck},
            \end{aligned}
        \end{equation}
        where:
        \begin{equation}
        \boxed{
            \ln \Tilde{\theta}_{dc} = \psi\left( \gamma_{dck} \right) - \psi\left( \sum_{k=1}^{K} \gamma_{dck} \right),
        }
            \label{eq:log_tilde_theta}
        \end{equation}
        and \(\psi(.)\) is the digamma function.
        
    \subsection{\texorpdfstring{\(\mathbb{E}_{q} \left[ \ln p(\bm{\theta} | \mathbf{y}, \bm{\alpha}) \right]\)}{}}
        \begin{equation}
            \begin{aligned}
                \mathbb{E}_{q} \left[ \ln p(\bm{\theta} | \mathbf{y}, \bm{\alpha}) \right] & = \sum_{d=1}^{M} \sum_{c=1}^{C} \sum_{l=1}^{L} q(y_{dcl} = 1; \eta_{dcl}) \int q(\bm{\theta}_{dc}; \bm{\gamma}_{dc}) \ln p(\bm{\theta}_{dc} | \bm{\alpha}_{l}) \dd{\bm{\theta}_{dcl}} \\
                & = \sum_{d=1}^{M} \sum_{c=1}^{C} \sum_{l=1}^{L} \eta_{dcl} \int \mathrm{Dirichlet}(\bm{\theta}_{dc}; \bm{\gamma}_{dc}) \ln \mathrm{Dirichlet}(\bm{\theta}_{dc} | \bm{\alpha}_{l}) \dd{\theta_{dcl}}.
            \end{aligned}
        \end{equation}
        Note that the cross-entropy between 2 Dirichlet distributions can be expressed as:
		\begin{equation}
    		\begin{aligned}[b]
    		\mathcal{H} \left[\mathrm{Dir} \left(\mathbf{x}; \bm{\alpha}_{0}\right), \mathrm{Dir} \left(\mathbf{x}; \bm{\alpha}_{1} \right) \right] & = - \mathbb{E}_{\mathrm{Dir} \left(\mathbf{x}; \bm{\alpha}_{0} \right)} \left[ \ln \mathrm{Dir} \left(\mathbf{x}; \bm{\alpha}_{1} \right) \right] \\
    		& = - \mathbb{E}_{\mathrm{Dir} \left(\mathbf{x}; \bm{\alpha}_{0} \right)}  \left[ - \ln B(\bm{\alpha_{1}}) + \sum_{k=1}^{K} (\alpha_{1k} - 1) \ln x_{k} \right] \\
    		& = \ln B(\bm{\alpha}_{1}) - \sum_{k=1}^{K} (\alpha_{1k} - 1) \left[ \psi(\alpha_{0k}) - \psi \left( \sum_{k^{\prime}=1}^{K} \alpha_{0k^{\prime}} \right) \right],
    		\end{aligned}
		\end{equation}
		where:
		\begin{equation}
			\ln B(\bm{\alpha}_{1}) = \sum_{k=1}^{K} \ln\Gamma \left( \alpha_{1k} \right) - \ln\Gamma \left( \sum_{j=1}^{K} \alpha_{1j} \right).
			\label{eq:log_dirichlet_normalized_const}
		\end{equation}
		Hence:
		\begin{equation}
			\begin{aligned}[b]
                \mathbb{E}_{q} \left[\ln p(\bm{\theta} \vert \mathbf{y}, \bm{\alpha}) \right] & = \sum_{d=1}^{M} \sum_{c=1}^{C} \sum_{l=1}^{L} \eta_{dcl} \left[ - \ln B(\bm{\alpha}_{l}) + \sum_{k=1}^{K} (\alpha_{lk} - 1) \ln \Tilde{\theta}_{dck} \right],
			\end{aligned}
		\end{equation}
		where \(\ln \Tilde{\theta}_{dck}\) is defined in Eq.~\eqref{eq:log_tilde_theta}.
	
	\subsection{\texorpdfstring{\(\mathbb{E}_{q} \left[\ln p(\mathbf{y} \vert \bm{\phi}) \right] \)}{}}
		\begin{equation}
			\begin{aligned}[b]
				\mathbb{E}_{q} \left[\ln p(\mathbf{y} \vert \bm{\phi}) \right] & = \sum_{d=1}^{M} \sum_{c=1}^{C} \sum_{l=1}^{L} q(y_{dcl} = 1; \eta_{dcl}) \int q(\bm{\phi}_{d}; \bm{\lambda}_{d}) \ln p(y_{dcl} = 1 \vert \phi_{dl}) \dd{\phi_{dl}} \\
				& = \sum_{d=1}^{M} \sum_{c=1}^{C} \sum_{l=1}^{L} \eta_{dcl} \int \mathrm{Dirichlet} (\bm{\phi}_{d}; \bm{\lambda}_{d}) \ln \phi_{dl}  \dd{\phi_{dl}} \\
				& = \sum_{d=1}^{M} \sum_{c=1}^{C} \sum_{l=1}^{L} \eta_{dcl} \ln \Tilde{\phi}_{dl},
			\end{aligned}
		\end{equation}
		where:
		\begin{equation}
		\boxed{
		    \ln \Tilde{\phi}_{dl} = \psi \left( \lambda_{dl} \right) - \psi \left( \sum_{j=1}^{K} \lambda_{dl} \right)
		}
		    \label{eq:log_tilde_phi}
		\end{equation}
	
	\subsection{\texorpdfstring{\(\mathbb{E}_{q} \left[\ln p(\bm{\phi} \vert \bm{\delta}) \right] \)}{}}
		\begin{equation}
			\begin{aligned}[b]
				\mathbb{E}_{q} \left[\ln p(\bm{\phi} \vert \bm{\delta}) \right] & = \sum_{d=1}^{M} \int q(\bm{\phi}_{d}; \bm{\lambda}_{d}) \ln p(\bm{\phi}_{d} \vert \bm{\delta}) \, d\bm{\phi}_{d} \\
				& = \sum_{d=1}^{M} \int \mathrm{Dirichlet}_{L} (\bm{\phi}_{d}; \bm{\lambda}_{d}) \ln \mathrm{Dirichlet}_{L} (\bm{\phi}_{d} \vert \bm{\delta}) \, d\bm{\phi}_{d} \\
				& = \sum_{d=1}^{M} - \ln B(\bm{\delta}) + \sum_{l=1}^{L} (\delta_{l} - 1) \ln \Tilde{\phi}_{dl},
			\end{aligned}
		\end{equation}
		where \(\ln \Tilde{\phi}_{dl}\) is defined in Eq.~\eqref{eq:log_tilde_phi}.
	
	\subsection{\texorpdfstring{\(\mathbb{E}_{q} \left[\ln q(\mathbf{z})\right] \)}{}}
		\begin{equation}
			\begin{aligned}[b]
				\mathbb{E}_{q} \left[\ln q(\mathbf{z})\right] & = \sum_{d=1}^{M} \sum_{c=1}^{C} \sum_{n=1}^{N} \sum_{k=1}^{K} r_{dcnk} \ln r_{dcnk}.
			\end{aligned}
		\end{equation}
	
	\subsection{\texorpdfstring{\(\mathbb{E}_{q} \left[\ln q(\bm{\theta})\right] \)}{}}
		\begin{equation}
			\begin{aligned}[b]
				\mathbb{E}_{q} \left[\ln q(\bm{\theta})\right] & = \sum_{d=1}^{M} \sum_{c=1}^{C} - \ln B(\bm{\gamma}_{dc}) + \sum_{j=1}^{K} (\gamma_{dck} - 1) \ln \Tilde{\theta}_{dck},
			\end{aligned}
		\end{equation}
		where \(\ln \Tilde{\theta}_{dck}\) is defined in Eq.~\eqref{eq:log_tilde_theta}.
	
	\subsection{\texorpdfstring{\(\mathbb{E}_{q} \left[\ln q(\mathbf{y})\right] \)}{}}
		\begin{equation}
			\mathbb{E}_{q} \left[\ln q(\mathbf{y})\right] = \sum_{d=1}^{M} \sum_{c=1}^{C} \sum_{l=1}^{L} \eta_{dcl} \ln \eta_{dcl}.
		\end{equation}
	
	\subsection{\texorpdfstring{\(\mathbb{E}_{q} \left[\ln q(\bm{\phi})\right] \)}{}}
		\begin{equation}
			\begin{aligned}[b]
				\mathbb{E}_{q} \left[\ln q(\bm{\phi})\right] & = \sum_{d=1}^{M} - \ln B(\bm{\lambda}_{d}) + \sum_{l=1}^{L} (\lambda_{dl} - 1) \ln \Tilde{\phi}_{dl},
			\end{aligned}
		\end{equation}
		where \(\ln \Tilde{\phi}_{dl}\) is defined in Eq.~\eqref{eq:log_tilde_phi}.

\section{Optimisation of the lower-bound}
\label{apdx:elbo_optimisation}
    \subsection{Variational categorical for \texorpdfstring{\(\mathbf{z}\)}{z}}
        The terms in the lower-bound that relates to \(r_{dcnk}\) are:
        \begin{equation}
            \begin{aligned}[b]
                \mathsf{L} & = -\frac{1}{2} r_{dcnk} \left[ D \ln(2\pi) + \ln |\bm{\Sigma}_{k}| + (\mathbf{x}_{dcn} - \bm{\mu}_{k})^{\top} \bm{\Sigma}_{k}^{-1} (\mathbf{x}_{dcn} - \bm{\mu}_{k})  \right] + r_{dcnk} \ln \Tilde{\theta}_{dck} \\
                & \qquad - r_{dcnk} \ln r_{dcnk} + \zeta \left( \sum_{k=1}^{K} r_{dcnk} - 1 \right),
            \end{aligned}
        \end{equation}
        where \(\ln \Tilde{\theta}_{dck}\) is defined in Eq.~\eqref{eq:log_tilde_theta}, and \(\zeta\) is the Lagrange multiplier due to the assumption that \(\mathbf{r}_{dcn}\) is the parameter of a categorical distribution, which requires:
        \begin{equation}
            \sum_{k=1}^{K} r_{dcnk} = 1.
        \end{equation}
        
        Taking the derivative w.r.t. \(r_{dcnk}\) gives:
        \begin{equation}
            \begin{aligned}[b]
                \pdv{\mathsf{L}}{r_{dcnk}} & = -\frac{1}{2} \left[ D \ln(2\pi) + \ln |\bm{\Sigma}_{k}| + (\mathbf{x}_{dcn} - \bm{\mu}_{k})^{\top} \bm{\Sigma}_{k}^{-1} (\mathbf{x}_{dcn} - \bm{\mu}_{k})  \right] \\
                & \quad + \ln \Tilde{\theta}_{dck} - \ln r_{dcnk} - 1 + \zeta
            \end{aligned}
        \end{equation}
        
        Setting the derivative to zero yields the maximizing value of the variational parameter \(r_{dcnk}\) as:
        \begin{equation}
            \boxed{
                r_{dcnk} \propto \exp \left\{ \ln \Tilde{\theta}_{dck} -\frac{1}{2} \left[ D \ln(2\pi) + \ln |\bm{\Sigma}_{k}| + (\mathbf{x}_{dcn} - \bm{\mu}_{k})^{\top} \bm{\Sigma}_{k}^{-1} (\mathbf{x}_{dcn} - \bm{\mu}_{k})  \right] \right\}.
            }
            \label{eq:r}
        \end{equation}
        
    \subsection{Variational Dirichlet for \texorpdfstring{\(\bm{\theta}\)}{theta}}
        The lower-bound isolating the terms for \(\gamma_{dck}\) is written as:
        \begin{equation}
            \begin{aligned}[b]
                \mathsf{L} & = \sum_{n=1}^{N} \sum_{k=1}^{K} r_{dcnk} \ln \Tilde{\theta}_{dck} + \sum_{l=1}^{L} \eta_{dcl} \sum_{k=1}^{K} (\alpha_{lk} - 1) \ln \Tilde{\theta}_{dck} - \ln B(\bm{\gamma}_{dc}) \\
                & \qquad + \sum_{k=1}^{K} (\gamma_{dck} - 1) \ln \Tilde{\theta}_{dck} \\
                & = - \ln B(\bm{\gamma}_{dc}) + \sum_{k=1}^{K} \ln \Tilde{\theta}_{dck} \left[ \sum_{n=1}^{N} r_{dcnk} + \sum_{l=1}^{L} \eta_{dcl} (\alpha_{lk} - 1) + \gamma_{dck} - 1 \right],
            \end{aligned}
        \end{equation}
        where \(\ln \Tilde{\theta}_{dck}\) is defined in Eq.~\eqref{eq:log_tilde_theta}.
        
        Taking derivative w.r.t. \(\gamma_{dck}\) gives:
        \begin{equation}
            \begin{aligned}[b]
                \pdv{\mathsf{L}}{\gamma_{dck}} & = \Psi(\gamma_{dck}) \left[\sum_{n=1}^{N} r_{dcnk} + \sum_{l=1}^{L} \eta_{dcl} (\alpha_{lk} - 1) -\gamma_{dck} + 1 \right] \\
    			& \quad - \Psi \left( \sum_{j=1}^{K} \gamma_{dcj} \right) \sum_{j=1}^{K} \left[\sum_{n=1}^{N} r_{dcnj} + \sum_{l=1}^{L} \eta_{dcl} (\alpha_{lj} - 1) -\gamma_{dcj} + 1 \right].
            \end{aligned}
        \end{equation}
        Setting the derivative to zero and solve for \(\gamma_{dck}\) yields:
    		\begin{equation}
    		\boxed{
    			\gamma_{dck} = 1 + \sum_{n=1}^{N} r_{dcnk} + \sum_{l=1}^{L} \eta_{dcl} (\alpha_{lk} - 1).
    		}
    		    \label{eq:gamma}
    		\end{equation}
    
    \subsection{Variational categorical for \texorpdfstring{\(\mathbf{y}\)}{y}}
        Note that the \(L\)-dimensional vector \(\bm{\eta}_{dc}\) is the parameter of a categorical distribution for \(\mathbf{y}_{dc}\), it satisfies the following constrain:
		\begin{equation}
			\sum_{l=1}^{L} \eta_{dcl} = 1.
		\end{equation}
		The Lagrangian can be expressed as:
		\begin{equation}
			\begin{aligned}[b]
				\mathsf{L}[\mathbf{y}_{dc}] & = \sum_{l=1}^{L} \eta_{dcl} \left[ - \ln B(\bm{\alpha}_{l}) + \sum_{k=1}^{K} (\alpha_{lk} - 1) \ln \Tilde{\theta}_{dck} \right] \\
				& \quad + \sum_{l=1}^{L} \eta_{dcl} \ln \Tilde{\phi}_{dl} - \sum_{l=1}^{L} \eta_{dcl} \ln \eta_{dcl} + \xi \left(\sum_{l=1}^{L} \eta_{dcl} - 1\right),
			\end{aligned}
		\end{equation}
		where \(\xi\) is the Lagrange multiplier, \(\ln \Tilde{\theta}_{dck}\) is defined in Eq.~\eqref{eq:log_tilde_theta}, and \(\ln \Tilde{\phi}_{dl}\) is defined in Eq.~\eqref{eq:log_tilde_phi}.
		
		Taking the derivative w.r.t. \(\eta_{dcl}\) gives:
		\begin{equation}
			\pdv{\mathsf{L}}{\eta_{dcl}} = - \ln B(\bm{\alpha}_{l}) + \sum_{k=1}^{K} (\alpha_{lk} - 1) \ln \Tilde{\theta}_{dck} + \psi \left( \lambda_{dl} \right) - \psi \left( \sum_{j=1}^{K} \lambda_{dl} \right) - \ln \eta_{dcl} - 1 + \xi.
		\end{equation}
		Setting the derivative to zero and solve for \(\eta_{dcl}\) yields:
		\begin{equation}
		\boxed{
			\eta_{dcl} \propto \exp \left[ \ln \Tilde{\phi}_{dl} - \ln B(\bm{\alpha}_{l}) + \sum_{k=1}^{K} (\alpha_{lk} - 1) \ln \Tilde{\theta}_{dck} \right].
		}
		    \label{eq:eta}
		\end{equation}
	
	\subsection{Variational Dirchlet for \texorpdfstring{\(\phi\)}{phi}}
	    The lower-bound isolating the terms for \(\lambda_{dl}\) is written as:
	    \begin{equation}
			\begin{aligned}[b]
				\mathsf{L} & = \sum_{c=1}^{C} \sum_{l=1}^{L} \eta_{dcl} \ln \Tilde{\phi}_{dl} + \sum_{l=1}^{L} (\delta_{l} - 1) \ln \Tilde{\phi}_{dl} + \ln B(\bm{\lambda}_{d}) - \sum_{l=1}^{L} (\lambda_{dl} - 1) \ln \Tilde{\phi}_{dl} \\
				& = \ln B(\bm{\lambda}_{d}) + \sum_{l=1}^{L} \ln \Tilde{\phi}_{dl} \left( \delta_{l} - \lambda_{dl} + \sum_{c=1}^{C} \eta_{dcl} \right),
			\end{aligned}
		\end{equation}
		where \(\ln \Tilde{\phi}_{dl}\) is defined in Eq.~\eqref{eq:log_tilde_phi}.
		
		Taking derivative w.r.t. \(\lambda_{dl}\) gives:
		\begin{equation}
			\begin{aligned}[b]
				\frac{\partial \mathsf{L}}{\partial \lambda_{dl}} & = \Psi(\lambda_{dl}) \left( \delta_{l} - \lambda_{dl} + \sum_{c=1}^{C} \eta_{dcl} \right) - \Psi \left(\sum_{j=1}^{L} \lambda_{dj}\right) \sum_{l=1}^{L} \left( \delta_{l} - \lambda_{dl} + \sum_{c=1}^{C} \eta_{dcl} \right).
			\end{aligned}
		\end{equation}
		
		Setting to zero and solving for \(\lambda_{dl}\) gives:
		\begin{equation}
		\boxed{
			\lambda_{dl} = \delta_{l} + \sum_{c=1}^{C} \eta_{dcl}.
		}
		    \label{eq:lambda}
		\end{equation}
	
	\subsection{Maximum likelihood for \texorpdfstring{\(\bm{\mu}\) and \(\bm{\Sigma}\)}{means and covariance matrices}}
	    The lower-bound isolating the terms for \(\bm{\mu}_{k}\) and \(\bm{\Sigma}_{k}\) is written as:
	    \begin{equation}
	        \mathsf{L} = - \frac{1}{2}\sum_{d=1}^{M} \sum_{c=1}^{C} \sum_{n=1}^{N} r_{dcnk} \left[ D \ln(2\pi) + \ln |\bm{\Sigma}_{k}| + (\mathbf{x}_{dcn} - \bm{\mu}_{k})^{\top} \bm{\Sigma}_{k}^{-1} (\mathbf{x}_{dcn} - \bm{\mu}_{k})  \right].
	    \end{equation}
	    Taking derivative w.r.t. \(\bm{\mu}_{k}\) and \(\bm{\Sigma}_{k}\) gives:
	    \begin{equation}
	        \begin{dcases}
	            \pdv{\mathsf{L}}{\bm{\mu}_{k}} & = \sum_{d=1}^{M} \sum_{c=1}^{C} \sum_{n=1}^{N} r_{dcnk} \bm{\Sigma}_{k}^{-1} (\mathbf{x}_{dcn} - \bm{\mu}_{k}) \\
	            \pdv{\mathsf{L}}{\bm{\Sigma}_{k}} & = - \frac{1}{2} \sum_{d=1}^{M} \sum_{c=1}^{C} \sum_{n=1}^{N} r_{dcnk} \left[ \bm{\Sigma}_{k}^{-1} - \bm{\Sigma}_{k}^{-1} (\mathbf{x}_{dcn} - \bm{\mu}_{k}) (\mathbf{x}_{dcn} - \bm{\mu}_{k})^{\top} \bm{\Sigma}_{k}^{-1} \right].
	        \end{dcases}
	    \end{equation}
	    Setting the derivative to zero yields the maximizing values at:
	    \begin{equation}
    		\boxed{
    		    \begin{aligned}[b]
        			\bm{\mu}_{k} & = \frac{1}{\sum_{d=1}^{M} N_{dk}} \sum_{d=1}^{M} \sum_{c=1}^{C} \sum_{n=1}^{N} r_{dcnk} \mathbf{x}_{dcn} \\
	                \bm{\Sigma}_{k} & = \frac{1}{\sum_{d=1}^{M} N_{dk}} \sum_{d=1}^{M} \sum_{c=1}^{C} \sum_{n=1}^{N} r_{dcnk} (\mathbf{x}_{dcn} - \bm{\mu}_{k}) (\mathbf{x}_{dcn} - \bm{\mu}_{k})^{\top},
    			\end{aligned}
    		}
    	    \label{eq:m_step}
    	\end{equation}
	    where:
	    \begin{equation}
	        N_{dk} = \sum_{c=1}^{C} \sum_{n=1}^{N} r_{dcnk}.
	    \end{equation}
	    
	    Note that the inference results for image-themes \(\{\bm{\mu}_{k}, \bm{\Sigma}_{k}\}_{k=1}^{K}\) in \eqref{eq:m_step} is very similar to the result of EM algorithm derived for a Gaussian mixture model~\cite[Chapter 9]{bishop2006pattern}. The result is, consequently, often suffered from the singularity issue happened in the MLE for a Gaussian mixture model. The issue is due to one of the Gaussian components collapses (or overfit) to a single data point, resulting in a zero covariance matrix. In the implementation, we add a small value (about \(10^{-6}\)) diagonal matrix to the covariance matrices to avoid this problem.

    \subsection{MLE for Dirichlet parameter \texorpdfstring{\(\bm{\alpha}\)}{alpha}}
    \label{apdx:alpha}
        The terms in ELBO which contains \(\bm{\alpha}\) are:
        \begin{equation}
            \begin{aligned}[b]
                \mathsf{L}[\bm{\alpha}] & = \sum_{d=1}^{M} \sum_{c=1}^{C} \sum_{l=1}^{L} \eta_{dcl} \left[ \ln\Gamma\left(\sum_{k=1}^{K} \alpha_{lk}\right) - \sum_{k=1}^{K} \ln\Gamma\left(\alpha_{lk}\right) + \sum_{k=1}^{K} (\alpha_{lk} - 1) \ln \Tilde{\theta}_{dck} \right].
            \end{aligned}
        \end{equation}
        
        Taking the derivative w.r.t. \(\alpha_{lk}\) gives:
        \begin{equation}
            \begin{aligned}
                \frac{\partial \mathsf{L}}{\partial \alpha_{lk}} & = g_{lk} =  M \left[ \psi \left(\sum_{k=1}^{K} \alpha_{lk}\right) - \psi\left(\alpha_{lk}\right) \right] \sum_{c=1}^{C} \eta_{dcl}  + \sum_{d=1}^{M} \sum_{c=1}^{C} \eta_{dcl} \ln \Tilde{\theta}_{dck}.
            \end{aligned}
        \end{equation}
        
        The Hessian matrix can be calculated as:
        \begin{equation}
            \begin{aligned}
                \frac{\partial^{2} \mathsf{L}}{\partial \alpha_{lk} \partial\alpha_{lj}} & = \underbrace{M \left[ \sum_{c=1}^{C} \eta_{dcl} \right] \Psi \left(\sum_{k=1}^{K} \alpha_{lk}\right)}_{u} \, \underbrace{- M \left[ \sum_{c=1}^{C} \eta_{dcl} \right] \mathbbm{1}[k = j] \, \Psi\left(\alpha_{lk}\right)}_{q_{ljk}}.
            \end{aligned}
        \end{equation}
        
        According to \cite{minka2000estimating}, Newton-Raphson method can be used to infer \(\bm{\alpha}_{l}\) as:
        \begin{align}
            \bm{\alpha}_{l} & \gets \bm{\alpha}_{l} - \mathbf{H}_{l}^{-1} \mathbf{g}_{l} \\
            \mathbf{H}_{l}^{-1} & = \mathbf{Q}_{l}^{-1} - \frac{\mathbf{Q}_{l}^{-1} \mathbf{11}^{\top} \mathbf{Q}_{l}^{-1}}{1 / u + \mathbf{1}^{\top} \mathbf{Q}_{l}^{-1} \mathbf{1}} \\
            \left(\mathbf{H}_{l}^{-1} \mathbf{g}_{l} \right)_{l} & = \frac{g_{lk} - b_{l}}{q_{lkk}} \label{eq:inverse_hessian_gradient},
        \end{align}
        where:
        \begin{equation}
            \begin{aligned}
                    b_{l} & = \frac{\mathbf{1}^{\top} \mathbf{Q}_{l}^{-1} \mathbf{g}_{l}}{1 / u + \mathbf{1}^{\top} \mathbf{Q}_{l}^{-1} \mathbf{1}} = \frac{\sum_{j} g_{lj} q_{ljj}}{1 / u + \sum_{j} 1 / q_{ljj}}.
            \end{aligned}
        \end{equation}
	\section{Learning algorithm}
\label{apdx:learning_algorithm}

    \begin{algorithm}[ht]
        \caption{Online continuous LDCC}
        \label{alg:online_ldcc}
        \begin{algorithmic}[1]
            \Statex \textbf{require} Scalar hyper-parameters: e-step stopping criteria \(\Delta\lambda\), learning rate parameters \(\tau_{0}, \tau_{1}\), and symmetric Dirichlet prior parameter \(\delta\)
            \Procedure{Training}{} 
            \State Initialise \(\{\bm{\mu}_{k}, \bm{\Sigma}_{k} \}_{k=1}^{K}\) and \(\{ \bm{\alpha}_{l} \}_{l=1}^{L}\)
            \For{\(d = 1, M\)}
                \State \(\bm{\lambda}, \bm{\eta}, \bm{\gamma}, \mathbf{r} \gets\) \Call{E-step}{$\mathbf{x}_{d}, \Delta\lambda$} \Comment{E-step}
                \State Calculate \say{local} image-theme \(\{\Tilde{\bm{\mu}}, \Tilde{\bm{\Sigma}} \}\) \Comment{M-step - Eq.~\eqref{eq:m_step}}
                \State Calculate the inverse of the Hessian times the gradient \(\mathbf{H}^{-1} \mathbf{g}\) \Comment{Eq.~\eqref{eq:inverse_hessian_gradient}}
                \State Update learning rate: \(\rho_{d} = (\tau_{0} + d)^{-\tau_{1}}\)
                \State \(\bm{\mu} \gets (1 - \rho_{d}) \bm{\mu} + \rho_{d} \Tilde{\bm{\mu}} \quad\) \Comment{Eq.~\ref{eq:online_update}}
                \State \(\bm{\Sigma} \gets (1 - \rho_{d}) \bm{\Sigma} + \rho_{d} \Tilde{\bm{\Sigma}} \quad\)
                \State \( \bm{\alpha} \gets \bm{\alpha} - \rho_{d} \mathbf{H}^{-1} \mathbf{g} \)
            \EndFor
            \State \textbf{return} \(\bm{\mu}, \bm{\Sigma}, \bm{\alpha}\)
            \EndProcedure
            
            \Statex
            
            \Procedure{E-step}{$\mathbf{x}, \Delta\lambda$}
                \State Initialise \(\mathbf{r}, \bm{\gamma}, \bm{\eta}, \bm{\lambda}\)
                \Repeat
                    \State calculate un-normalised \(r_{cnk}\), where \(n \in \{1, \ldots, N\}, k \in \{1, \ldots, K\}\) \Comment{Eq.~\eqref{eq:r}}
                    \State normalize \(\mathbf{r}_{cn}\) such that \(\sum_{k=1}^{K} r_{cnk} = 1\)
                    \State calculate \(\gamma_{ck}\) \Comment{Eq.~\eqref{eq:gamma}}
                    \State calculate \(\eta_{cl}\), where: \(l \in \{1, \ldots, L\}\) \Comment{Eq.~\eqref{eq:eta}}
                    \State normalize \(\bm{\eta}_{c}\) such that \(\sum_{l=1}^{L} \eta_{cl} = 1\)
                    \State calculate \(\lambda_{l}\) \Comment{Eq.~\eqref{eq:lambda}}
                \Until{\(\frac{1}{L}~| \text{change in } \bm{\lambda} | < \Delta\lambda\) }
                \State \textbf{return} \(\bm{\lambda}, \bm{\eta}, \bm{\gamma}, \mathbf{r}\)
            \EndProcedure
        \end{algorithmic}
    \end{algorithm}
\end{document}